\documentclass[letterpaper]{article} 
\usepackage{aaai25}  
\usepackage{times}  
\usepackage{helvet}  
\usepackage{courier}  
\usepackage[hyphens]{url}  
\usepackage{graphicx} 
\usepackage{natbib}  
\usepackage{caption} 
\usepackage{algorithm}
\usepackage{algorithmic}

\usepackage{newfloat}
\usepackage{listings}
\DeclareCaptionStyle{ruled}{labelfont=normalfont,labelsep=colon,strut=off} 
\floatstyle{ruled}
\newfloat{listing}{tb}{lst}{}
\floatname{listing}{Listing}
\usepackage{xcolor}

\title{Speech Is Not Enough: Interpreting Nonverbal Indicators of Common Knowledge and Engagement}
\author{
   Derek Palmer\textsuperscript{\rm 1},
   Yifan Zhu\textsuperscript{\rm 1},
   Kenneth Lai\textsuperscript{\rm 1},
    Hannah VanderHoeven\textsuperscript{\rm 2},
    Mariah Bradford\textsuperscript{\rm 2},
    Ibrahim Khebour\textsuperscript{\rm 2},
    Carlos Mabrey\textsuperscript{\rm 2},
    Jack Fitzgerald\textsuperscript{\rm 2},
   Nikhil Krishnaswamy\textsuperscript{\rm 2},\\
    Martha Palmer\textsuperscript{\rm 3},
    James Pustejovsky\textsuperscript{\rm 1}
}
\affiliations{

    \textsuperscript{\rm 1}Brandeis University\\ 
    \textsuperscript{\rm 2}Colorado State University\\ 
    \textsuperscript{\rm 3}University of Colorado Boulder\\
    $\{$derek.palmer@colorado.edu, jamesp@brandeis.edu$\}$
}

\usepackage{bibentry}

\begin{document}

\maketitle

\begin{abstract}

Our goal is to develop an AI Partner that can provide support for group problem solving and social dynamics. In multi-party working group environments, multimodal analytics is crucial for identifying non-verbal interactions of group members. In conjunction with their verbal participation, this creates an holistic understanding of collaboration and engagement that provides necessary context for the AI Partner.
In this demo, we illustrate our present capabilities at detecting and tracking nonverbal behavior in student task-oriented interactions in the classroom, and the implications for tracking common ground and engagement.
\end{abstract}

\section{Introduction}
Our goal is developing an AI partner that can provide beneficial information or suggestions to groups of collaborators in real time. Essential to this process is an accurate interpretation of two dimensions of the AI partner's environment: the working group's knowledge  of the topic, and the current social dynamics of the group.  Multi-modal analysis offers unique analysis of vital non-verbal cues \cite{deyCSCL2023}. Also, the more complex and novel the environment, the less reliable automatic speech recognition is. Multi-channel input is crucial for useful input to AI partners. A demo video is at \url{https://youtu.be/WzajCzOYggg}. 

\paragraph{The Task} Our Institute for Student-AI Teaming (iSAT), aims to develop an AI Partner that can intervene positively in collaborative problem solving groups of students \cite{iSAT-AIMagazine}. Student group productivity can be heavily influenced by social dynamics \cite{moulder2022assessing}. Social dynamics can be positive  or negative based on the behavior and level of engagement of each individual member \cite{adams2022marginality}. Social cohesion is positive social dynamics manifesting as high levels of engagement for all group members culminating in constructive progress towards the group's goal. Negative social cohesion can be either low levels of engagement of the group with little progress towards the goal or unconstructive interactions.  Automatic speech recognition is a vital part of identifying both positive and negative social situations. However, as the amount of topics, participants, and background noise expands, ASR accuracy decreases \cite{cao2023comparative}, increasing reliance on multimodal analysis, especially with speaker cohorts such as children with minimal training data. 
Also, many non-verbal interactions are simply inaccessible to ASR and critical to capture via multi-modal analysis. 

Tracking gestures such as pointing can be indispensable in building real-time understanding of a group's common knowledge \cite{khebour-etal-2024-common,tu2024dense}. Tracking each individual's posture over time, in particular leaning in or leaning out, is a powerful indicator of a group's engagement level \cite{adams2022marginality}.  Joint visual attention is critical to contextualize both common knowledge and group engagement. All together these modalities help pinpoint intervention opportunities and avoid unconstructive interruptions \cite{Jie-JIA-2023}. 
 We illustrate multi-modal analysis in both  dimensions: 1) a knowledge support analysis with our Fibonacci weights exercise; 2) contrasting levels of engagement only observable via multi-modal analysis for our simulated classroom environment.

\section{Our Setup}
The physical task space consists of a table with task-relevant objects on it and 3 participants seated around it. The task is recorded using an Azure Kinect RGBD camera, and an MXL AC-404 ProCon microphone. The Kinect automatically tracks 32 joints per body, covering head, torso, and limbs, returning 3 position and 4 orientation values for each. 

\textbf{Gaze detection} uses the direction of participants' noses as a proxy. Using the joints of bodies extracted by the Azure Kinect SDK, we take the average position of both ear joints, which results in a point roughly behind the nose, and use a vector connecting this point and the nose joint to indicate gaze direction. We extend this vector (purple) into 3D space to see which objects participants' gazes are landing on.

{\bf Posture detection} determines the participants' positions (left, middle, right) using the $x$-coordinate of the pelvis of each body. Each participant's position and orientation information is then flattened. The vectors are stacked and 
then input into a two-layer feedforward neural network. We train three such models, one for each participant position.

\textbf{Gesture recognition} primarily concerns pointing detection. We use a lightweight 2-stage method from \citet{vanderhoeven2023robust,vanderhoeven2024point} with features extracted from the video using MediaPipe~\cite{lugaresi2019mediapipe}. First we detect if a gesture is in the ``stroke'' phase (following \citet{kendon1997gesture}) and then classify the gesture's shape. As with gaze, to infer the target objects of a pointing gesture, we calculate a ``pointing frustum''~\cite{kranstedt2006deixis} from the extended digit into 3D space (blue and orange in the video). Objects intersecting this frustum are considered ``selected'' (highlighted green).

\textbf{Real-time object detection} in the video is performed using a FasterRCNN ResNet-50-FPN model \cite{lin2017feature}, initialized using the default ResNet-50-FPN weights from TorchVision, then trained over annotations of object bounding boxes for 10 epochs using SGD with learning rate $1e-3$, momentum $9e-1$, and weight decay $5e-4$. Batch size was 32 and input size was 3$\times$416$\times$416.


\subsection{Video Content}
\begin{figure}
    \centering
    \includegraphics[width=.8\linewidth]{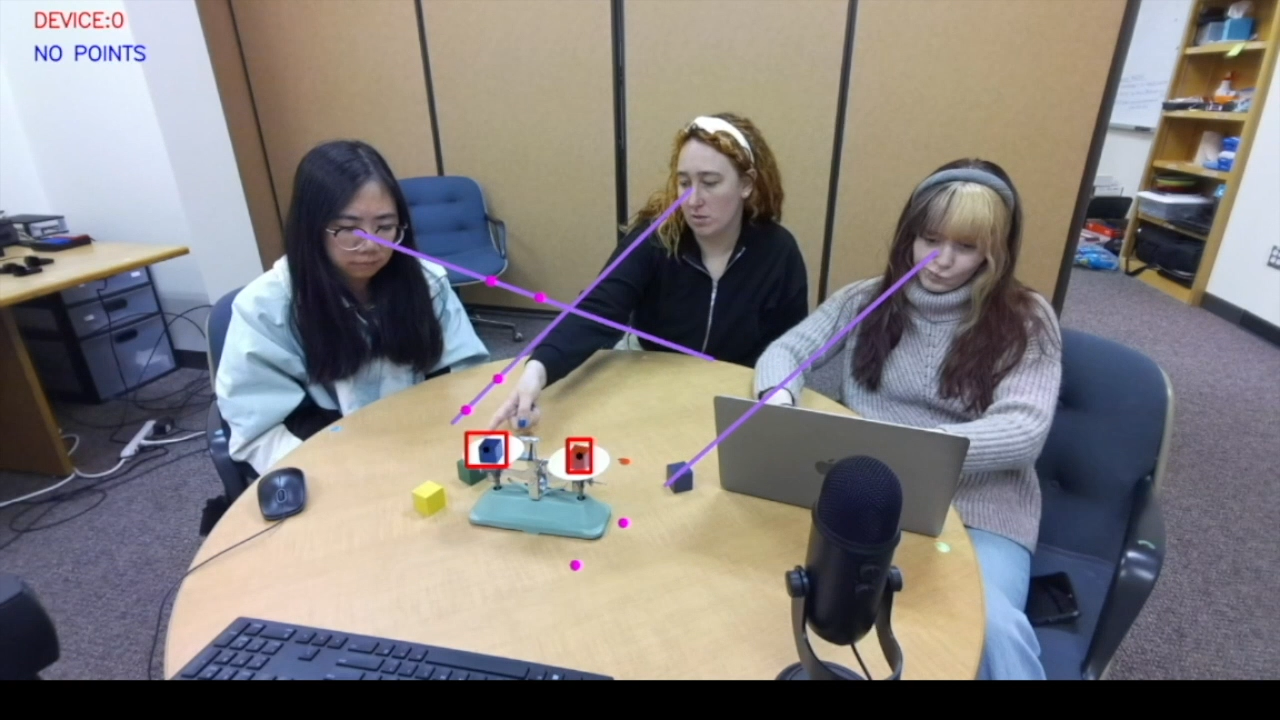}
    \caption{Object detection in Weights Task}
    \label{fig:scenario1}
\end{figure}

\begin{figure}
    \centering
    \includegraphics[width=.8\linewidth]{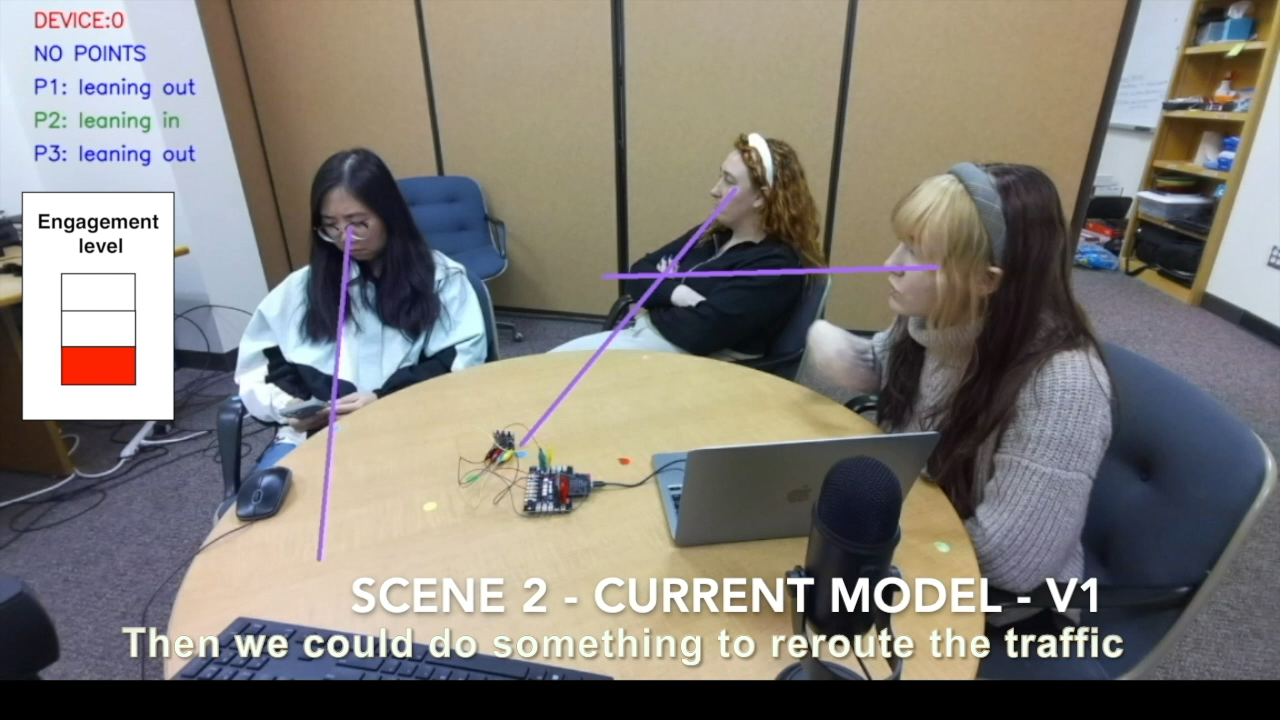} \\
    \includegraphics[width=.8\linewidth]{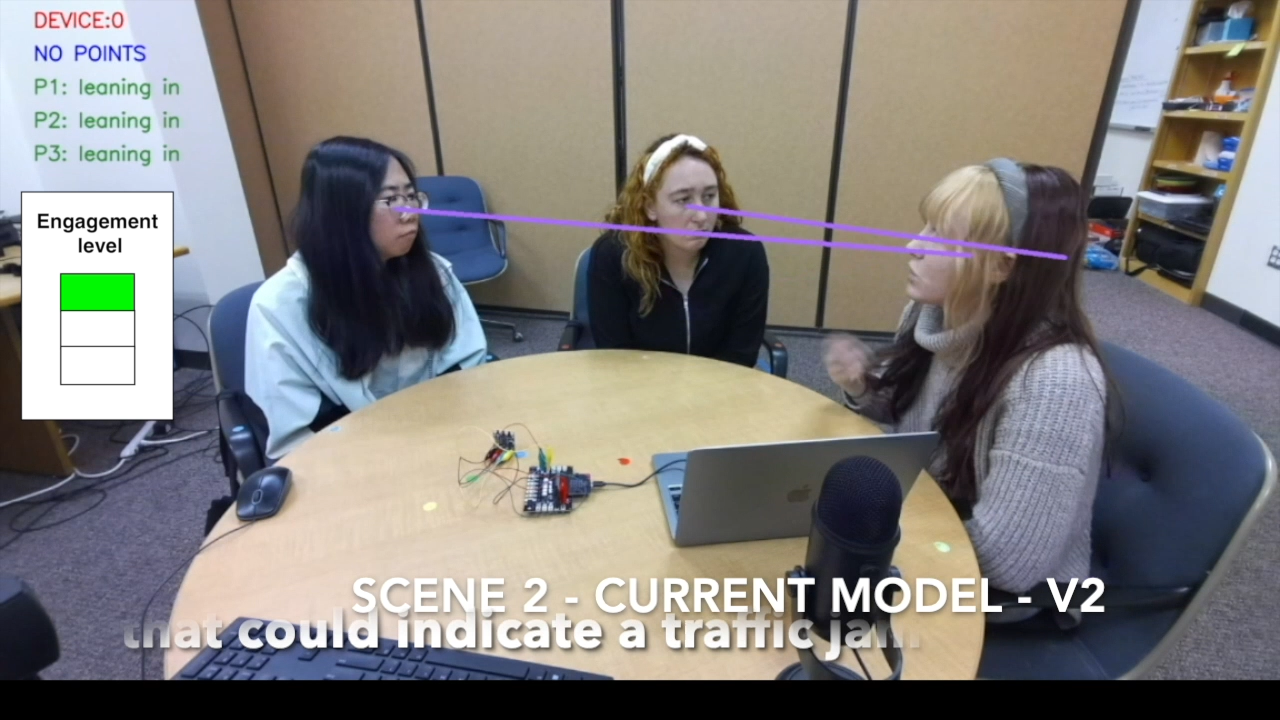}
    \caption{Contrasting engagement levels in simulated project planning}
    \label{fig:scenario2}
\end{figure}

\textbf{Our two scenarios} (Figs.~\ref{fig:scenario1} and \ref{fig:scenario2}) prioritize different aspects of multimodal information processing: knowledge support often includes the evaluation of specific objects and classroom discussions do not. 
The context for knowledge support is equally dependent on joint visual attention, gesture and domain specific object detection. For social cohesion, joint visual attention to speakers and posture are more valuable. 

\textbf{Scenario 1: Fibonacci Weights Task}  To better evaluate the accuracy and utility of object detection, we developed a situated collaborative task wherein participants infer the weights of a set of differently weighted blocks with the use of a balance scale~\cite{khebour2024text}.  The increases in weight adhere to the Fibonacci series. A series of lab experiments provided video data for annotation and training purposes (see Fig.~\ref{fig:scenario1}).  The expectation is that object detection provides valuable input for AI Partners offering knowledge support.  With minimal amounts of training data we can port to similar new objects specific to new domains.



\textbf{Scenario 2: Simulated Classroom Project Planning} The simulated classroom content we are demoing is a project planning scenario from Lesson 4 of the Sensor Immersion Curriculum Unit developed by the SchoolWide Labs project \cite{swl2021}. In the lesson each student comes into the group as the sole expert on one of three specific sensor types. 
The students answer factual questions about the capabilities of their sensors and brainstorm about potential beneficial projects involving multiple sensors in real-world environments. An AI Partner \cite{Jie-JIA-2023} guides and supports these conversations, via a chat window. 

Deciding which problem to address at any given time is a compromise between what can be detected, theories of behavior based on behavioral/cognitive models, and defined lesson goals. 
Non-verbal Multimodal analysis is primarily useful for determining whether social dynamics are positive or negative, relative to the specific domain under investigation.
Our AI partner has several knowledge support and social cohesion states it attempts to recognize, as defined and vetted by educational researchers \cite{JIA-CHI}. An example of a state, Dominated Discussion, is the detection of a single group member talking for 30 seconds or more. Our demo illustrates the necessity of using multi-modal analysis to accurately identify whether the dominated discussion results in disengaged participants (Fig. \ref{fig:scenario2}).


\section{Conclusion and Future Work} 
We have demonstrated the importance of integrating nonverbal behavior recognition into the modeling and interpretation of multi-party dialogues when there is an intervention objective. With an eye on portability, our next focus for object detection will be devices common to most settings such as tablets, laptops and phones.  We expect that our approach to modeling social cohesion will port to any AI partner included in a working group of 3 or more, given a document summarizing the specific topic and a task specific  model of engagement \cite{MooreSurgical2016,ClosedDoors2009}. This could include business, government, health, or education settings, such as board meetings, working task forces, training exercises, etc. An opportunity also exists for decreasing the time intensive labor of video annotation tasks for qualitative research purposes.


\section{Acknowledgements}  We are extremely grateful to the brilliant actors in our demo, Rui Zhang, Sierra Rose and Brooklyn Clines. This material is based in part upon work supported by the National Science Foundation (NSF) under subcontracts to Colorado State University and Brandeis University on award DRL 2019805 (Institute for Student-AI Teaming), and by Other Transaction award HR00112490377 from the U.S. Defense Advanced Research Projects Agency (DARPA) Friction for Accountability in Conversational Transactions (FACT) program. Approved for public release, distribution unlimited. Views expressed herein do not reflect the policy or position of the National Science Foundation, the Department of Defense, or the U.S. Government.


\bibliography{aaai25}

\end{document}